\relax

\documentclass[letterpaper]{article}
\usepackage{uai2019}
\usepackage[margin=1in]{geometry}
\usepackage{times}

\usepackage{graphicx}
\usepackage[square,sort]{natbib}

\usepackage[utf8]{inputenc} 
\usepackage[T1]{fontenc}    
\usepackage{hyperref}       
\usepackage{amsfonts}       
\usepackage{nicefrac}       
\usepackage{microtype}      
\usepackage{amsmath,amsthm,verbatim,amssymb,amscd,mathrsfs}
\usepackage{dsfont}
\usepackage{enumitem}
\usepackage{xcolor}
\usepackage{algorithm,algorithmic}
\usepackage{wrapfig,subcaption}
\usepackage{url}

\renewcommand{\paragraph}[1]{\vspace{0.40ex}\noindent\textbf{#1}}

\definecolor{mypurple}{RGB}{128,0,128}

\newcommand{\bx}{\mathbf{x}}
\newcommand{\by}{y}
\newcommand{\bz}{\mathbf{z}}

\DeclareMathOperator{\dct}{DCT}
\DeclareMathOperator{\idct}{IDCT}

\hypersetup{draft}

\begin{document}
%

\title{Low Frequency Adversarial Perturbation}

%


\author{Chuan Guo, Jared S. Frank, Kilian Q. Weinberger\\
Cornell University, Ithaca, NY 14853, USA\\
\{cg563, jsf239, kqw4\}@cornell.edu
}

\maketitle

\begin{abstract}
Adversarial images aim to change a target model's decision by minimally perturbing a target image. In the black-box setting, the absence of gradient information often renders this search problem costly in terms of query complexity. In this paper we propose to restrict the search for adversarial images to a low frequency domain. This approach is readily compatible with many existing black-box attack frameworks and consistently reduces their query cost by 2 to 4 times. Further, we can circumvent image transformation defenses even when both the model and the defense strategy are unknown. Finally, we demonstrate the efficacy of this technique by fooling the Google Cloud Vision platform with an unprecedented low number of model queries.
\end{abstract}

\section{INTRODUCTION}
As machine learning models enjoy widespread adoption, their security becomes a relevant topic for consideration. Recent studies have shown that existing methods lack robustness against imperceptible changes to the input \citep{biggio2013evasion, szegedy2013intriguing}, and many deployed computer vision and speech recognition systems have been compromised \citep{liu2016delving, melis2017robot, cisse2017houdini, carlini2018audio, ilyas2018blackbox}. This presents a realistic security threat in critical applications such as autonomous driving, where an adversary may manipulate road signs to cause control malfunction while remaining hidden to the naked eye \citep{evtimov2017robust}.

Most existing attack algorithms, both white-box \citep{szegedy2013intriguing, dezfooli2016deepfool, carlini2017towards} and black-box \citep{chen2017zoo, brendel2017decision, tu2018autozoom, ilyas2018blackbox}, function by searching the full space of possible perturbations to find noise patterns that alter the behavior of convolutional filters. In this high dimensional space many solutions exist and search algorithms tend to almost exclusively result in high frequency solutions, \emph{i.e.} small pixel-wise perturbations dispersed across an image. White-box attacks can be guided by gradient information and tend to have low query complexity (as low as 10 gradients on ResNet/ImageNet).
In contrast, black-box attacks do not enjoy such benefits. For example, the search for successful ResNet/ImageNet attacks still requires on the order of $10^4-10^5$ queries.

\begin{figure}[t]
\centering
\includegraphics[width=\columnwidth]{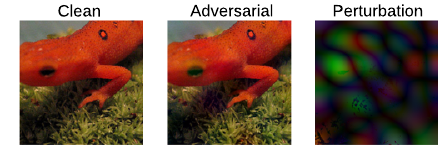}
\vspace{-1.5em}
\caption{A sample low frequency adversarial image produced by black-box attack.}
\label{fig:blackbox_sample}
\vspace{-1.5em}
\end{figure}

Motivated by these shortcomings, we propose a radical departure from the existing, high-frequency adversarial perturbation attacks and we explicitly restrict the search space of adversarial directions to the low frequency subspace. 
Constructing low frequency adversarial perturbation has several advantages: As black-box attacks generally require random sampling in the image space, its high-dimensionality causes the attack algorithm to sample many non-adversarial directions, resulting in a high query complexity on the order of the image dimensionality. In the low frequency subspace adversarial directions may occur in much higher density -- lowering query complexity significantly. Moreover, many successful defenses against black-box attacks rely on removing high frequency signal with a low-pass filter \citep{dziugaite2016study, xu2017feature, guo2017countering}, and operating in low frequency space promises to bypass these image transformation defenses.


In this paper we show that adversarial perturbations do indeed exist abundantly in a very low-dimensional low frequency subspace. We demonstrate that two popular black-box attacks -- the boundary attack \citep{brendel2017decision} and the natural evolution strategies (NES) attack \citep{ilyas2018blackbox} -- can be readily restricted to such a low frequency domain. Figure \ref{fig:blackbox_sample} shows a sample black-box adversarial image with low frequency perturbation produced by the boundary attack. Our experiments demonstrate that a dimensionality reduction to a mere $1/64$ of the original space still yields near-optimal adversarial perturbations. Experimental results confirm our conjectured benefits in the black-box setting:

1. The boundary attack with low frequency perturbation requires dramatically fewer model queries to find an adversarial image. The modified attack produces adversarial images with imperceptible change on ImageNet (ResNet-50~\citep{he2016residual}) after approximately 1000 median number of model queries -- a 4x reduction compared to vanilla boundary attack.

2. The NES attack enjoys significant improvement of query efficiency with this simple modification, resulting in a consistent 2x speed-up across all images. The median number of queries required for a \emph{targeted} black-box attack using low frequency NES is only around $12,000$.

3. Using low frequency perturbation circumvents image transformation defenses such as JPEG compression \citep{dziugaite2016study} and bit depth reduction \citep{xu2017feature}, which have not exhibited vulnerability to black-box attacks prior to our work.

4. Finally, we employ the low frequency boundary attack to fool the Google Cloud Vision platform with an unprecedented 1000 model queries --- demonstrating its cost effectiveness and real world applicability.

\section{BACKGROUND}
\label{sec:background}

In the study of adversarial examples in image classification, the goal of an attacker is to alter the model's prediction by adding an imperceptible perturbation to a natural image. Formally, for a given classification model $h$ and an image $\bx$ on which the model correctly predicts $\by = h(\bx)$, the adversary aims to find a perturbed image $\bx'$ that solves the following constrained optimization problem:
$$\min_{\bx'} \hspace{4pt} \delta(\bx, \bx') \hspace{4pt} \text{subject to} \hspace{4pt} h(\bx') \neq \by.$$
The function $\delta$ measures the perceptual difference between the original and adversarial images, and is often approximated by mean squared error (MSE), the Euclidean norm $\| \cdot \|_2$ or the max-norm $\| \cdot \|_\infty$. An attack is considered successful if the perturbed image is imperceptibly different, i.e., $\delta(\bx, \bx') \leq \rho$ for some small $\rho > 0$. This attack goal defines an \emph{untargeted attack}, since the attack goal is to alter the prediction on the perturbed image to any incorrect class $h(\bx') \neq \by$. In contrast, a \emph{targeted} attack aims to produce perturbed images that the model predicts as some pre-specified target class.

When constructing adversarial images, the attacker may have various degrees of knowledge about the model $h$, including the training data and/or procedure, model architecture, or even all of its parameters. The attack may also adaptively query $h$ on chosen inputs before producing the adversarial images and obtain gradients from $h$. These different threat models can be roughly categorized into \emph{white-box}, where the attacker has full knowledge about $h$ and how it is trained, or \emph{black-box}, where the attacker can only query $h$, and has limited knowledge about its architecture or training procedure.

\paragraph{White-box attacks.} When given access to the model entirely, the adversary may minimize the correct class prediction probability directly to cause misclassification \citep{goodfellow2015explaining, kurakin2016adversarial, carlini2017towards, madry2017towards}. For a given input $\bx$ and correct class $\by$, the adversary defines a loss function $\ell_{\by}(\bx')$ so that the loss value is low when $h(\bx') \neq \by$. One example of such $\ell$ is the margin loss
\begin{equation}
    \ell_{\by}(\bx') = \max\left(Z(\bx')_{\by} - \max_{y' \neq \by} Z(\bx')_{y'} + \kappa ,0\right)
    \label{eq:margin_loss}
\end{equation}
used in \citep{carlini2017towards}, where $Z$ is the logit output of the network. The loss diminishes to zero only if the logit of at least one class exceeds that of the correct class, $y$, by $\kappa$ or more.  
The adversary can then solve
$$\min_{\bx'} \ell_{\by}(\bx') + \lambda \delta(\bx, \bx')$$
with a suitable hyperparameter $\lambda$ to constrain the perturbation to be small while ensuring misclassification.

\paragraph{Black-box attacks.} In certain scenarios, the white-box threat model does not reflect the true capability of an attacker. For example, when attacking machine learning services such as Google Cloud Vision, the attacker only has access to a limited number of function calls against the target model, and does not have knowledge about the training data. \emph{Transfer-based attacks} \citep{papernot2017practical, liu2016delving, tramer2017ensemble} utilize a substitute model that the attacker trains to imitate the target model, and constructs adversarial examples on the substitute model using white-box attacks. For this attack to succeed, the target model must be similar to the substitute model and is trained on the same data distribution. \emph{Gradient estimation attacks} use techniques such as finite difference \citep{chen2017zoo, tu2018autozoom} and natural evolution strategies \citep{ilyas2018blackbox, ilyas2018prior} to estimate the gradient from input-output pairs, thus enabling gradient-based white-box attacks. This type of attack requires the model to output class scores or probabilities, and generally requires a number of model queries proportional to the image size. In contrast, \emph{decision-based attacks} \citep{brendel2017decision, ilyas2018blackbox} utilize only the discrete classification decisions from a model and is applicable in all scenarios, but is generally more difficult to execute.

\section{LOW FREQUENCY IMAGE SUBSPACE}
\label{sec:method}

The inherent query inefficiency of gradient estimation and decision-based attacks stems from the need to search over or randomly sample from the high-dimensional image space. Thus, their query complexity depends on the relative adversarial subspace dimensionality compared to the full image space. One way to improve these methods is to find a low-dimensional subspace that contains a high density of adversarial examples, which enables more efficient sampling of useful attack directions.

Methods in image compression, in particular the celebrated JPEG codec~\citep{wallace1991jpeg}, have long exploited the fact that most of the critical content-defining information in natural images live in the low end of the frequency spectrum, whereas high frequency signal is often associated with noise. It is therefore plausible to assume that CNNs are trained to respond especially to low-frequency patterns in order to extract class-specific signatures from images. Hence, we propose to target CNN based approaches by restricting the search space for adversarial directions to the low-frequency spectrum -- essentially targeting these class defining signatures directly.



\paragraph{Discrete cosine transform.}
\label{sec:dct}
The JPEG codec utilizes the \emph{discrete cosine transform} (DCT), which decomposes a signal into cosine wave components, to represent a natural image in frequency space. More precisely, given a 2D image $X \in \mathbb{R}^{d \times d}$, define basis functions
\begin{equation*}
\phi_d(i,j) = \cos \left[ \frac{\pi}{d} \left(i + \frac{1}{2}\right) j \right]
\end{equation*}
for $1 \leq i,j \leq d$. The DCT transform $V = \dct(X)$ is:
\begin{equation*}
V_{j_1,j_2} = N_{j_1} N_{j_2} \sum_{i_1=0}^{d-1} \sum_{i_2=0}^{d-1} X_{i_1,i_2} \phi_d(i_1,j_1) \phi_d(i_2,j_2),
\label{eq:dct}
\end{equation*}
where $N_j=\sqrt{\frac{1}{d}}$ if $j=0$ and $N_j=\sqrt{\frac{2}{d}}$ otherwise. 
Here, $N_{j_1}, N_{j_2}$ are normalization terms included to ensure the transformation is isometric, i.e. $\|X\|_2 = \|\dct(X)\|_2$. The entry $V_{i,j}$ corresponds to the magnitude of wave $\phi_d(i,j)$, with lower frequencies represented by lower $i,j$. Further, DCT is invertible, with inverse $X = \idct(V)$,
\begin{equation}
X_{i_1,i_2} = \sum_{j_1=0}^{d-1} \sum_{j_2=0}^{d-1} N_{j_1} N_{j_2} V_{j_1,j_2} \phi_d(i_1,j_1) \phi_d(i_2,j_2).
\label{eq:idct}
\end{equation}
For images containing multiple color channels, both DCT and IDCT can be applied channel-wise independently.

\paragraph{Sampling low frequency noise.}
In order to facilitate efficient search for attack directions in low frequency space, we need to be able to sample random perturbations confined to this subspace. We can achieve this with the inverse DCT transform by considering the top-left $rd \times rd$ entries of $V$ for some ratio parameter $r \in (0,1]$. These coefficients correspond to cosine waves with long periods, hence low frequency. Given any distribution $\mathcal{D}$ (e.g. uniform, or Gaussian) over $\mathbb{R}^{d \times d}$, we can sample a random matrix $\tilde{\eta} \in \mathbb{R}^{d \times d}$ in frequency space so that
\begin{equation*}
\tilde{\eta}_{i,j} =
\begin{cases}
x_{i,j} \sim \mathcal{D} & \text{if } 1 \leq i,j \leq rd \\
0 & \text{otherwise.}
\end{cases}
\end{equation*}
Using the inverse DCT mapping, the corresponding noise ``image'' in pixel space is defined by $\eta = \idct(\tilde{\eta})$. By definition, $\eta$ has non-zero cosine wave coefficients only in frequencies lower than $rd$. When the pixel space contains multiple color channels, we can sample each channel independently using the same strategy. We denote this distribution of low frequency noise as $\idct_r(\mathcal{D})$ and the sub-space as low frequency DCT (LF-DCT) space. 


\begin{figure}[t!]
\begin{subfigure}{0.48\columnwidth}
  \centering
  \includegraphics[width=\textwidth]{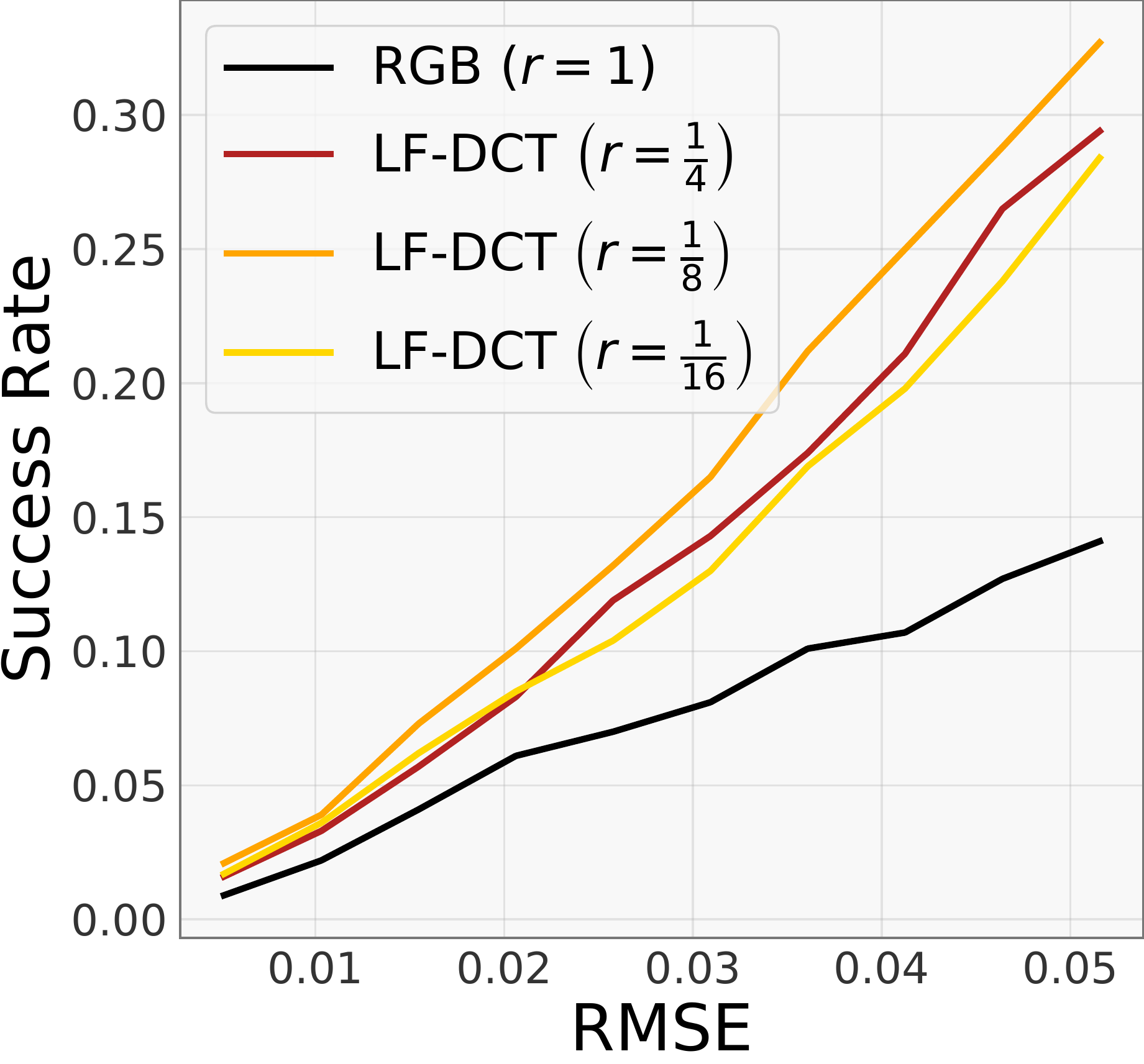}
\end{subfigure}
\begin{subfigure}{0.48\columnwidth}
  \centering
  \includegraphics[width=\textwidth]{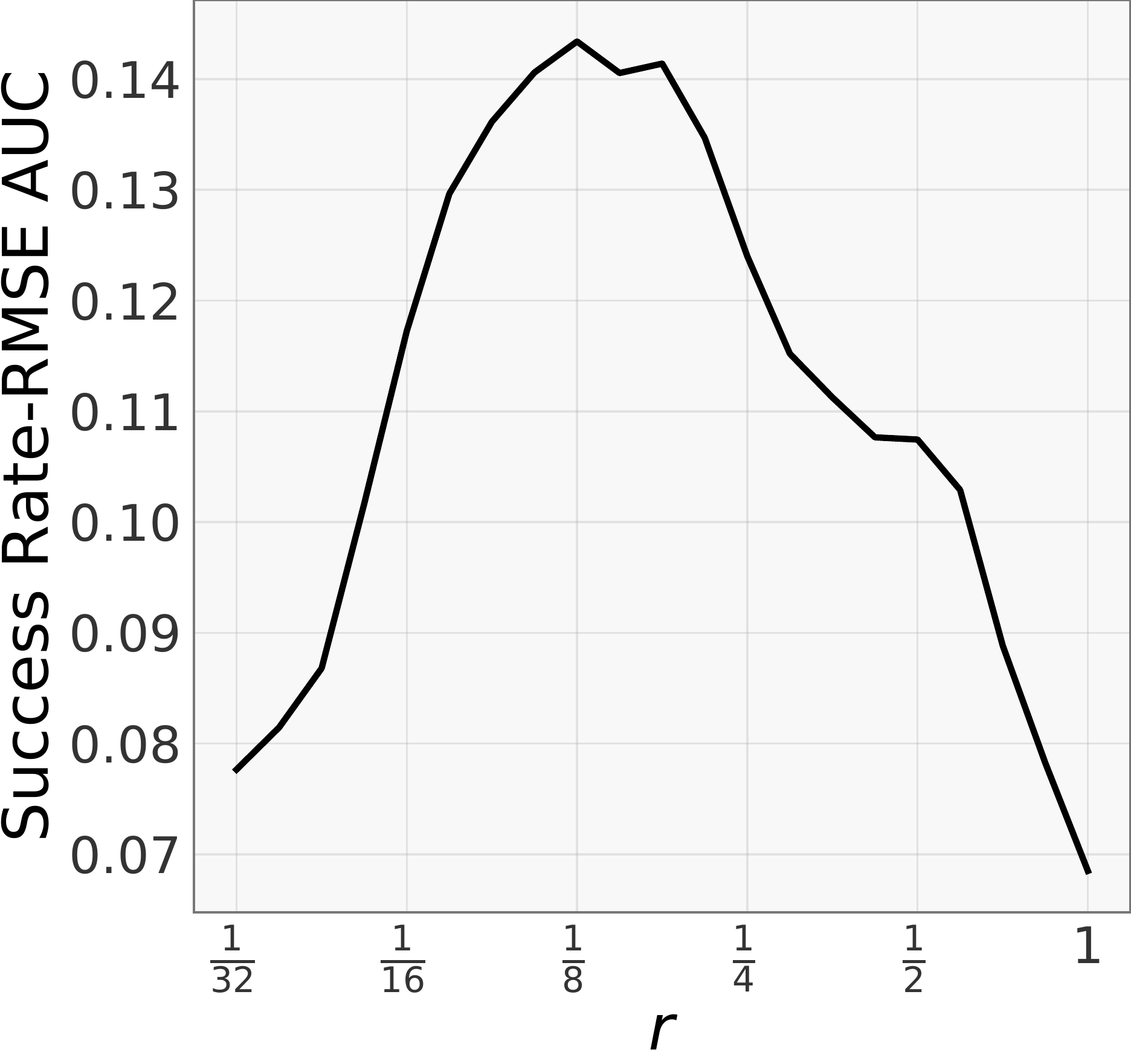}
\end{subfigure}
\vspace{-0.5em}
\caption{(Left) Comparison of (attack) success rate after perturbation by random spherical noise in RGB vs. LF-DCT space. Using low frequency noise improves the success rate dramatically. (Right) Area under the \emph{success rate}-$\rho$ curve. The highest density of adversarial images appears to lie around $r=1/8$.}
\label{fig:random_noise}
\vspace{-2ex}
\end{figure}

\paragraph{Low frequency noise success rate.}
We postulate that CNNs are more sensitive to changes in the LF-DCT subspace, hence admitting a higher density of adversarial perturbations. To empirically substantiate this hypothesis, we compare the success rate of random noise in RGB vs. LF-DCT space for a ResNet-50 architecture~\citep{he2016residual}. We sample the noise vector $\eta$ uniformly from the surface of the unit sphere of radius $\rho>0$ in the $rd\times rd$ LF-DCT space and project it back to RGB through the IDCT transform. For $r=1$ this procedure is identical to sampling directly on the surface of a unit sphere in pixel space, as IDCT is a linear, orthonormal transformation. 

Figure \ref{fig:random_noise} (left) shows these success rates as a function of the root mean squared error (RMSE $=\frac{\rho}{\sqrt{3}d}$) between the perturbed and original image, averaged over 1000 randomly chosen images from ImageNet~\citep{deng2009imagenet}. Several trends emerge: 1. As expected, the success rate increases with the magnitude of perturbation $\rho$ across all values of $r$; 2. There appears to be a sweet spot around $r=1/8$, which corresponds to a reduction of dimensionality by $1/64$; 3. The worst success rate is achieved with $r=1$, which corresponds to no dimensionality reduction (and is identical to sampling in the original RGB space). 

To further investigate the dimensionality ``sweet spot'', the right plot shows the area under the success rate vs. RMSE curve for various values of $r$, within a fixed range of $\rho\in[0,20]$. Here, a higher value corresponds to a faster increase in success rate with larger perturbation radius. In agreement with the left plot, the optimal frequency ratio is around $r=1/8$.

\subsection{Universality of low frequency subspace}

Results in the previous section support our hypothesis that restricting the search space to LF-DCT substantially increases the sample success rate of random adversarial directions. However, the dimensionality reduction does impose a restriction on the possible solutions of attack algorithms. To examine the effects of this limitation, we apply our low-frequency restriction to white-box attacks by projecting the gradient onto the LF-DCT space. 


\paragraph{Low frequency gradient descent.} 
Let $\ell_{y}$ denote the adversarial loss, e.g.  \autoref{eq:margin_loss}. For a given  $r \in (0, 1]$ and $v \in \mathbb{R}^{rd \times rd}$, define $V \in \mathbb{R}^{d \times d}$ by
\begin{equation*}
V_{i,j} =
\begin{cases}
v_{i,j} & \text{if } 1 \leq i,j \leq rd \\
0 & \text{otherwise,}
\end{cases}
\end{equation*}
The wave coefficient matrix $V$ contains $v$ as its submatrix and only includes frequencies lower than $rd$. The low frequency perturbation domain can then be parametrized as $\Delta = \idct(V)$. To optimize with gradient descent, let $\bar{\Delta}$ and $\bar{V}$ be vectorizations of $\Delta$ and $V$, i.e., $\bar{\Delta}_{i_1 * d + i_2} = \Delta_{i_1, i_2}$ and similarly for $\bar{V}$. From \autoref{eq:idct}, it is easy to see that each coordinate of $\bar{\Delta}$ is a linear function of $\bar{V}$, hence $\idct$ is a linear transformation, whose adjoint is precisely the linear transformation defined by $\dct$. For any vector $\bz$, its right-product with the Jacobian of $\idct$ is given by $J_{\idct} \cdot \bz = \dct(\bz)$. Thus we may apply the chain rule to compute
\begin{equation*}
\frac{\partial \ell}{\partial V} = \dct \left( \frac{\partial \ell}{\partial \Delta} \right), \frac{\partial \ell}{\partial v} = \left[\frac{\partial \ell}{\partial V}\right]_{1:rd,1:rd},
\end{equation*}
which is equivalent to applying DCT to the gradient and dropping the high frequency coefficients. 

\begin{figure}
\centering
\includegraphics[width=\columnwidth]{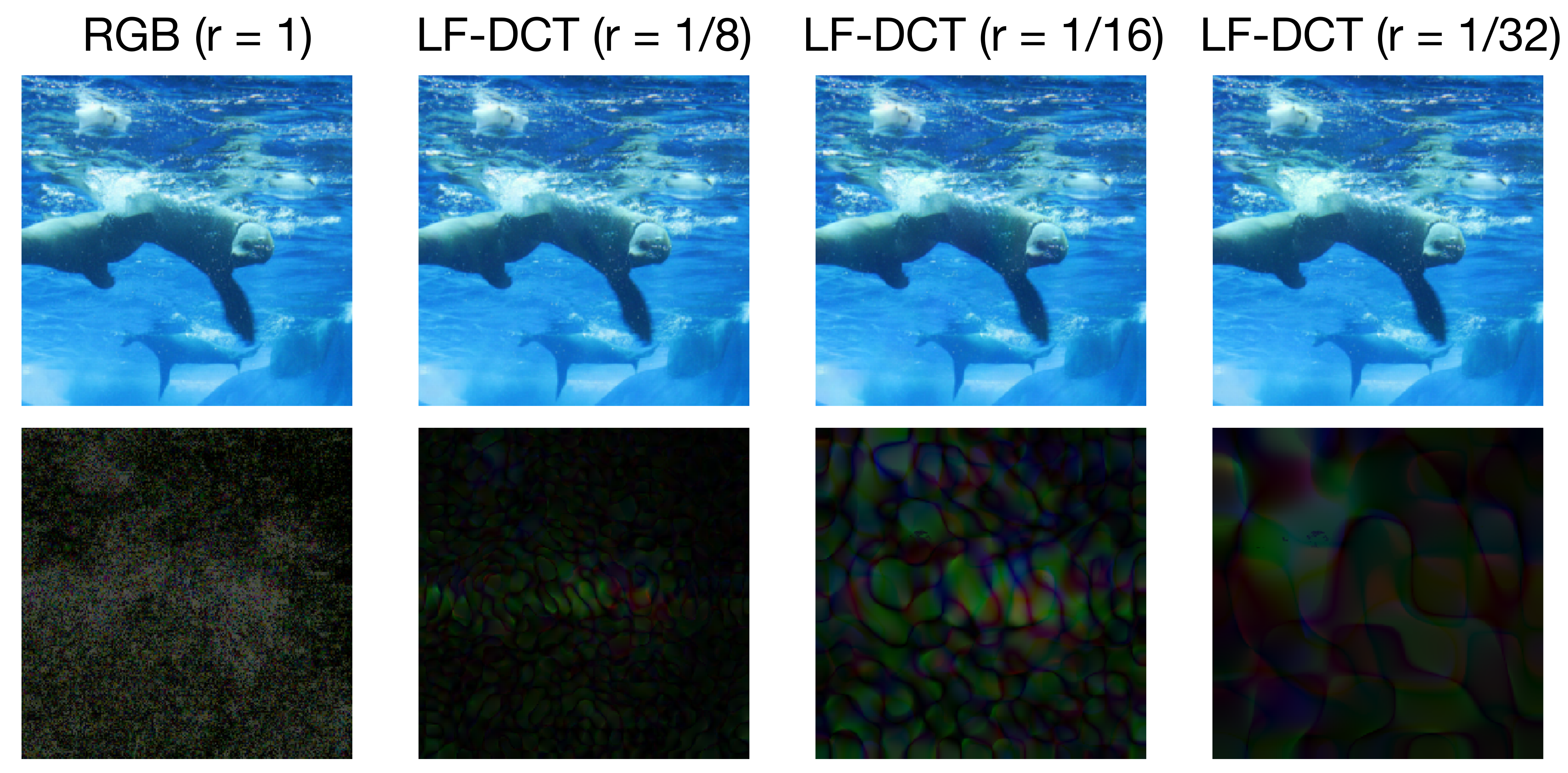}
\vspace{-3ex}
\caption{A sample image perturbed by the Carlini-Wagner attack using the full image space and low frequency space with different $r$. The adversarial perturbation (second row) has clearly different pattern across different frequency ranges.}
\label{fig:whitebox_sample}
\end{figure}

\begin{table}
\centering
\resizebox{\columnwidth}{!}{%
\begin{tabular}{cccc}
\hline
~& $d'$ & MSE & Success Rate (\%) \\
\hline
RGB ($r = 1$) & 150528 & $2.78 \times 10^{-5}$ & 100.0 \\
LF-DCT ($r = 1/8$) & 2352 & $6.94 \times 10^{-5}$ & 100.0 \\
LF-DCT ($r = 1/16$) & 588 & $1.61 \times 10^{-4}$ & 95.5 \\
LF-DCT ($r = 1/32$) & 147 & $1.56 \times 10^{-4}$ & 56.0 \\
\hline
\end{tabular}
}
\caption{Average MSE and accuracy after Carlini-Wagner attack with different frequency ratios $r$. $d' = 3 \times rd \times rd$ is the effective adversarial space dimensionality. At $r=1/8$, optimizing in the frequency space of dimensionality 2352 is as effective as optimizing in the full image space.}
\label{table:whitebox_mse}
\vspace{-1em}
\end{table}

\paragraph{Adversarial optimality in low frequency subspace.} 
Table \ref{table:whitebox_mse} shows average perturbation MSE and model accuracy after the Carlini-Wagner attack \citep{carlini2017towards} in low frequency space. The original attack in pixel space corresponds to $r = 1$. The images have three color channels and the effective subspace dimensionality is $d' = 3 \times rd \times rd$. For $r = 1/8$, the attack can achieve perfect (100\%) success rate, while the resulting MSE is only roughly 3 times larger --- despite that the search space dimensionality is only $1/64$ of the full image space. 
This result further supports that the density of adversarial examples is much higher in the low frequency domain, and that searching exclusively in this restricted subspace consistently yields near-optimal adversarial perturbations.
As expected, choosing a very small frequency ratio eventually impacts success rate, as the subspace dimensionality is too low to admit adversarial perturbations.
\autoref{fig:whitebox_sample} shows the resulting adversarial images and perturbations corresponding to frequency ratios $r$. All perturbations are imperceptible but when isolated (bottom row) reveal increasingly smooth  patterns as r decreases. 

\paragraph{Advantages of low frequency perturbation.} While the remainder of this paper focuses on the benefits of low frequency adversarial perturbation in the black-box setting, we highlight that there are advantages in the white-box setting as well. \citet{sharma2019low} showed that low frequency gradient-based attacks enjoy greater efficiency and can transfer significantly better to defended models. In particular, their attack is able to completely circumvent all of the top-placing defense entries at the NeurIPS 2017 competition. Furthermore, they observe that the benefit of low frequency perturbation is not merely due to dimensionality reduction --- perturbing exclusively the high frequency components does not give the same benefit.

\section{APPLICATION TO BLACK-BOX ATTACKS}
\label{sec:application}
\begin{figure}[t]
\centering
\includegraphics[width=\columnwidth]{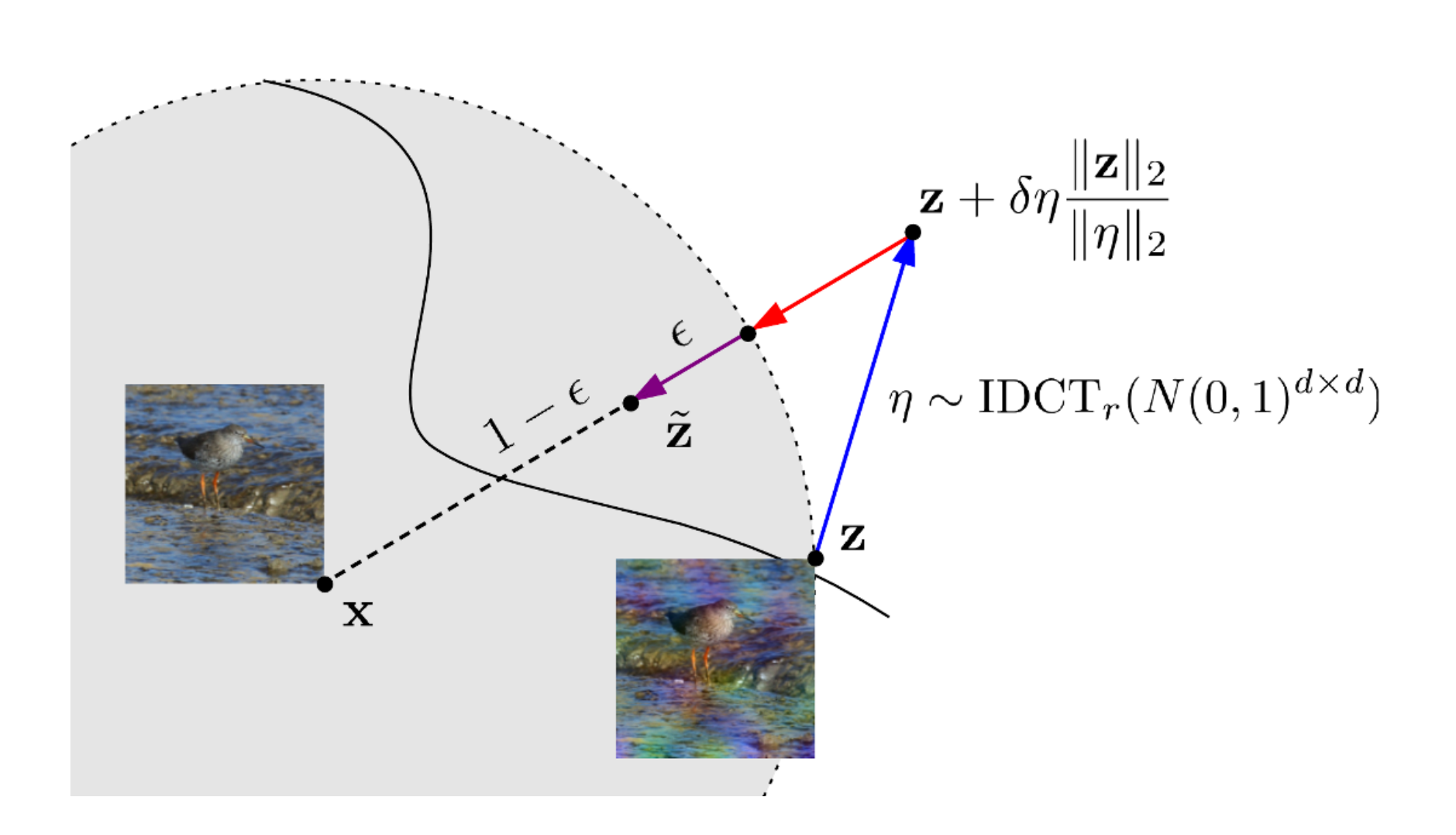}
\vspace{-1.5em}
\caption{Illustration of a single iteration of the low frequency boundary attack. Instead of sampling the noise matrix $\eta$ from $N(0,1)^{d \times d}$, we sample a low frequency noise matrix by applying IDCT to the Gaussian noise while removing high frequency components.
\label{fig:boundary}}
\vspace{-2ex}
\end{figure}

Many existing black-box attacks proceed by iteratively adding random noise to the current image and evaluating the model to determine the direction to move towards. Given our insights regarding the effectiveness of low frequency perturbations, we propose its use as a universal tool for improving the query efficiency of black-box attacks. We conduct case studies on the boundary attack \citep{brendel2017decision} and the NES attack \citep{ilyas2018blackbox} to demonstrate the efficacy and accessibility of our method.

\subsection{Case study: Boundary attack}

The boundary attack uses an iterative update rule to gradually move the adversarial image closer to the original image, maintaining that the image remains adversarial at each step. Starting from random noise, the algorithm samples a noise matrix $\eta \sim N(0,1)^{d \times d}$ at each iteration and adds it to the current iterate $\bz$ after appropriate scaling. This point is then projected onto the sphere of center $\bx$, the target image, and radius $\|\bz\|_2$ so that the next iterate never moves away from $\bx$. Finally, we contract towards $\bx$ by $\epsilon$, and the new iterate $\tilde{\bz}$ is accepted only if it remains adversarial. This guarantees that terminating the algorithm at any point still results in an adversarial image, but the perturbation magnitude reduces with each contraction step.

\paragraph{Modification.} To construct low frequency perturbation using the boundary attack, we constrain the noise matrix $\eta$ to be sampled from $\idct_r(N(0,1)^{d \times d})$ instead. Figure~\ref{fig:boundary} illustrates the modified attack. Sampling low frequency noise instead of Gaussian noise is particularly beneficial to the boundary attack: After adding the noise matrix $\eta$, if the iterate is not adversarial, the algorithm must re-sample a noise matrix and perform another model query. By restricting to the low frequency subspace, which has a larger density of adversarial directions, this step succeeds more often, speeding up convergence towards the target image. We term this variant of the boundary attack as \emph{low frequency boundary attack} (LF-BA) and the original boundary attack as RGB-BA.

\begin{figure*}[t!]
\centering
\includegraphics[width=0.49\textwidth]{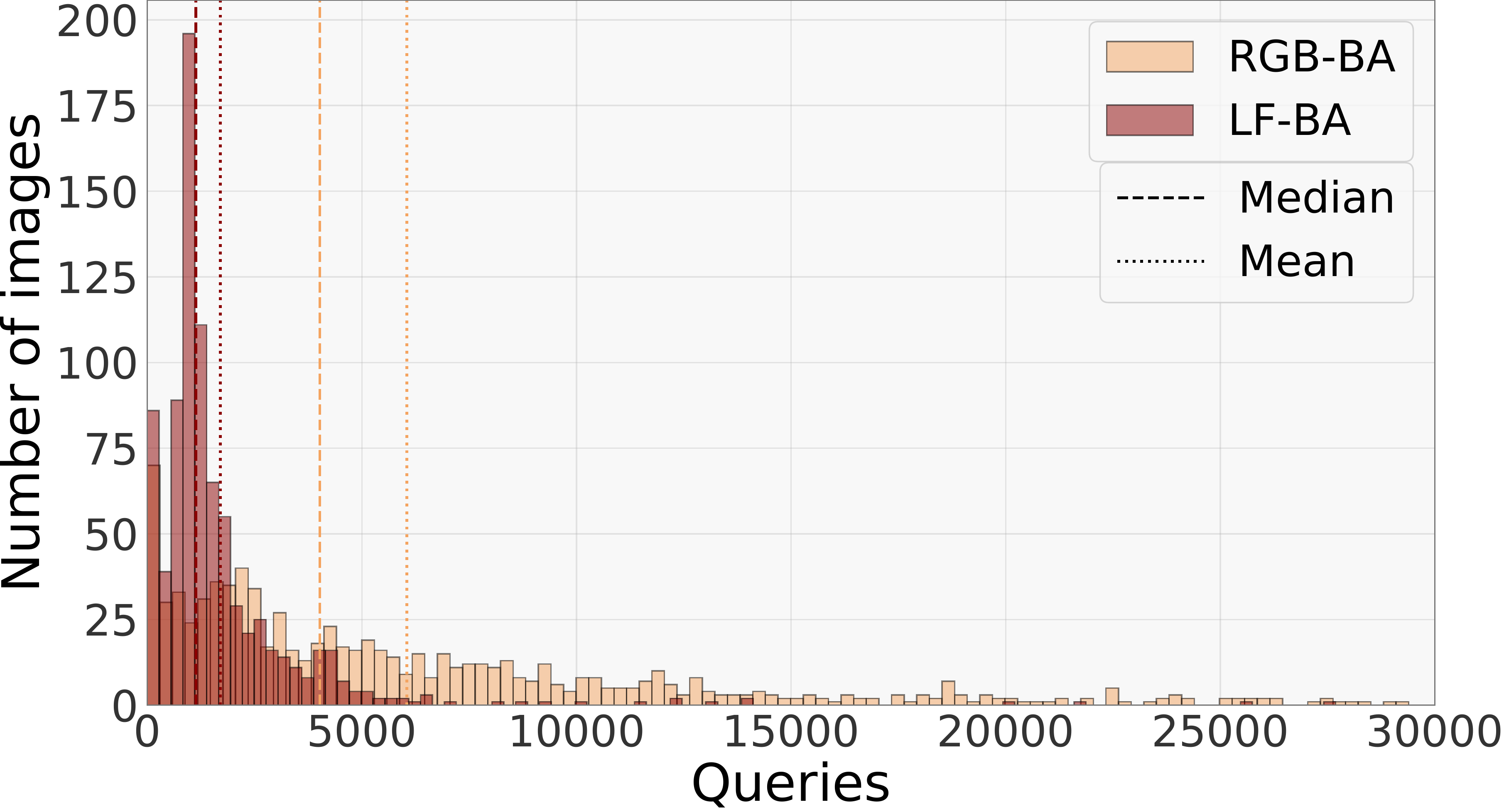}
\includegraphics[width=0.49\textwidth]{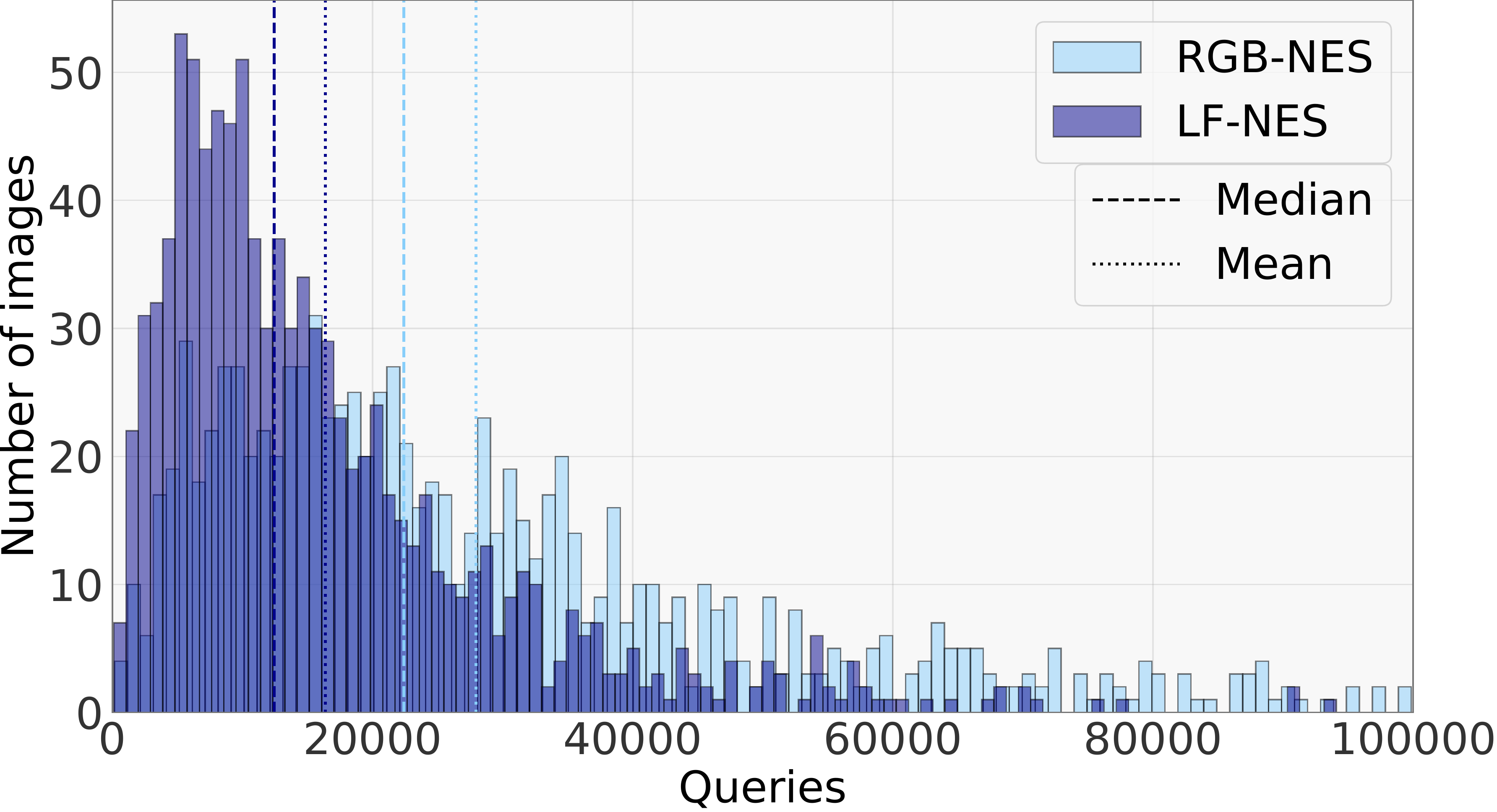}
\caption{Distribution of the number of queries required for a successful attack (defined as achieving a perturbation MSE of 0.001 or lower for RGB-BA/LF-BA). See text for details.} 
\label{fig:queries_comparison}
\vspace{-1em}
\end{figure*}

\paragraph{Hyperparameters.} The boundary attack has two hyperparameters: noise step size $\delta$ and contraction step size $\epsilon$. Both step sizes are adjusted based on the success rate of the past few candidates, \emph{i.e.}, if $\tilde{\bz}$ is accepted often, we can contract towards the target $\bx$ more aggressively by increasing $\epsilon$ and vice versa, and similarly for $\delta$. For the low frequency variant, we find that fixing $\delta$ to a large value is beneficial for speeding up convergence, while also reducing the number of model queries by half. For all experiments, we fix $\delta = 0.2$ and initialize $\epsilon = 0.01$.

Selecting the frequency ratio $r$ is more crucial. Different images may admit adversarial perturbations at different frequency ranges, and thus we would like the algorithm to automatically discover the right frequency on a per-image basis. We use Hyperband \citep{li2016hyperband}, a bandit-type algorithm for selecting hyperparameters, to optimize the frequency ratio $r$. We initialize Hyperband with multiple runs of the attack for every frequency ratio $r \in \{\frac{1}{4},\frac{1}{8},\frac{1}{16},\frac{1}{32}\}$. Repeatedly after $T'$ iterations, the least successful half of the parallel runs is terminated until one final frequency remains. This setting is continued until the total number of model queries reaches $T$.

\subsection{Case study: NES attack}

Natural evolution strategies (NES) \citep{wierstra2014natural} is a black-box optimization technique that has been recently proposed for its use in black-box attacks \citep{ilyas2018blackbox}. The attacker constructs the adversarial image $\bz$ by minimizing a continuous-valued adversarial loss $\ell$ returned by black-box query to the model. However, instead of minimizing $\ell$ directly, the NES attack minimizes the loss at all points near $\bz$. More precisely, we specify a search distribution $\mathcal{D}$ and minimize:
\begin{equation}
    \min_{\bz} \mathbb{E}_{\eta \sim \mathcal{D}}[\ell(\bz + \eta)] \hspace{4pt} \text{subject to} \hspace{4pt} d(\bx, \bz) \leq \rho,
    \label{eq:nes}
\end{equation}
where $\rho$ is some perceptibility threshold. When the search distribution $\mathcal{D}$ is chosen to be an isotropic Gaussian, i.e. $\mathcal{D} = N(0, \sigma^2)^{d \times d}$, the gradient of the objective function in \autoref{eq:nes} becomes
\begin{equation*}
    \nabla_{\bz} \mathbb{E}_{\eta \sim \mathcal{D}}[\ell(\bz + \eta)] = \frac{1}{\sigma^2} \mathbb{E}_{\eta \sim \mathcal{D}}[\ell(\bz + \eta) \cdot \eta].
\end{equation*}
Thus, \autoref{eq:nes} can be minimized with stochastic gradient descent by sampling a batch of noise vectors $\eta_1,\ldots,\eta_m \sim N(0, \sigma^2)^{d \times d}$ and computing the (mini-batch) stochastic gradient
\begin{equation}
    \nabla_{\bz} \mathbb{E}_{\eta \sim \mathcal{D}}[\ell(\bz + \eta)] \approx \frac{1}{m \sigma^2} \sum_{i=1}^m \ell(\bz + \eta_i) \cdot \eta_i.
    \label{eq:nes_update}
\end{equation}
One way to interpret this update rule is that the procedure pushes $\bz$ away from regions of low adversarial density --- directions $\eta_i$ for which $\ell(\bz + \eta_i)$ is high.
The perceptibility constraint can be enforced by projecting to the feasible region at every step. For this attack, the max-norm $\| \cdot \|_\infty$ is used as the perceptibility metric, hence the projection step reduces to clipping of each dimension in the adversarial perturbation to the range $[-\rho, \rho]$.

\paragraph{Modification.} The low frequency distribution defined in \autoref{sec:dct} can be readily incorporated into the NES attack. We replace the Gaussian search distribution with its low frequency version, i.e. we sample a batch of noise vectors $\eta_1,\ldots,\eta_m \sim \idct_r(N(0, \sigma^2)^{d \times d})$ instead. The stochastic gradient remains identical to \autoref{eq:nes_update}. Note that since each $\eta_i$ is low-frequency, this process results in a low frequency adversarial perturbation. We term the original NES attack using search distribution in pixel space as RGB-NES and the low frequency variant as \emph{low frequency NES} (LF-NES).


\paragraph{Hyperparameters.} The NES attack has two hyperparameters: $\rho$, which controls the perceptibility of adversarial perturbation, and $\sigma$, which controls the width of the search distribution. We set $\rho = 0.03$ to match the average $L_2$-norm of perturbations generated by RGB-BA/LF-BA, and set $\sigma = 0.001$ as suggested by the authors. Intriguingly, the frequency ratio $r$ is not very sensitive for LF-NES. Setting a single value of $r$ for all images is sufficiently effective, and we choose the same value of $r = 1/2$ in all of our experiments for simplicity.

\section{EMPIRICAL EVALUATION}
\label{sec:experiment}


We empirically validate our claims that black-box attacks in low frequency space possess the aforementioned desirable properties. For all experiments, we use the default PyTorch pretrained ResNet-50 model for RGB-BA/LF-BA and the Tensorflow-Slim pretrained ResNet-50 model\footnote{\url{https://github.com/tensorflow/models/tree/master/research/slim}} for RGB-NES/LF-NES.

Both RGB-BA and LF-BA use a 10 step binary search along the line joining the random initialization and the target image before starting the attack. Our implementation of the boundary attack in PyTorch has comparable performance to the official implementation by \citet{brendel2017decision} while being significantly faster. We use the official implementation of NES in Tensorflow and modify it to use low frequency search distribution. We release our code\footnote{\url{https://github.com/cg563/low-frequency-adversarial}} publicly for reproducibility.

\paragraph{Settings.} For experiments on ImageNet \citep{deng2009imagenet}, we evaluate both untargeted attack (RGB-BA/LF-BA) and targeted attack (RGB-NES/LF-NES) to a random class against a pretrained ResNet-50 \citep{he2016residual} model. Each test image is randomly selected from the validation set while ensuring correct prediction by the respective models. For RGB-BA/LF-BA, the adversary can only access the binary output of the model corresponding to whether the input is classified as the original label. For RGB-NES/LF-NES, the adversary can obtain the cross entropy loss of the model against the target label.

We limit the attack algorithm to $30,000$ queries for untargeted attack, corresponding to $30,000$ iterations for LF-BA and $15,000$ for RGB-BA\footnote{RGB-BA requires two model queries per iteration, one after the noise step and one after the contraction.}, and $100,000$ queries for targeted attack using RGB-NES/LF-NES. For LF-BA, we select the frequency ratio $r$ using Hyperband by halving the number of parallel runs every $T' = 500$ iterations.

\begin{figure}[t!]
\centering
\includegraphics[width=0.9\columnwidth]{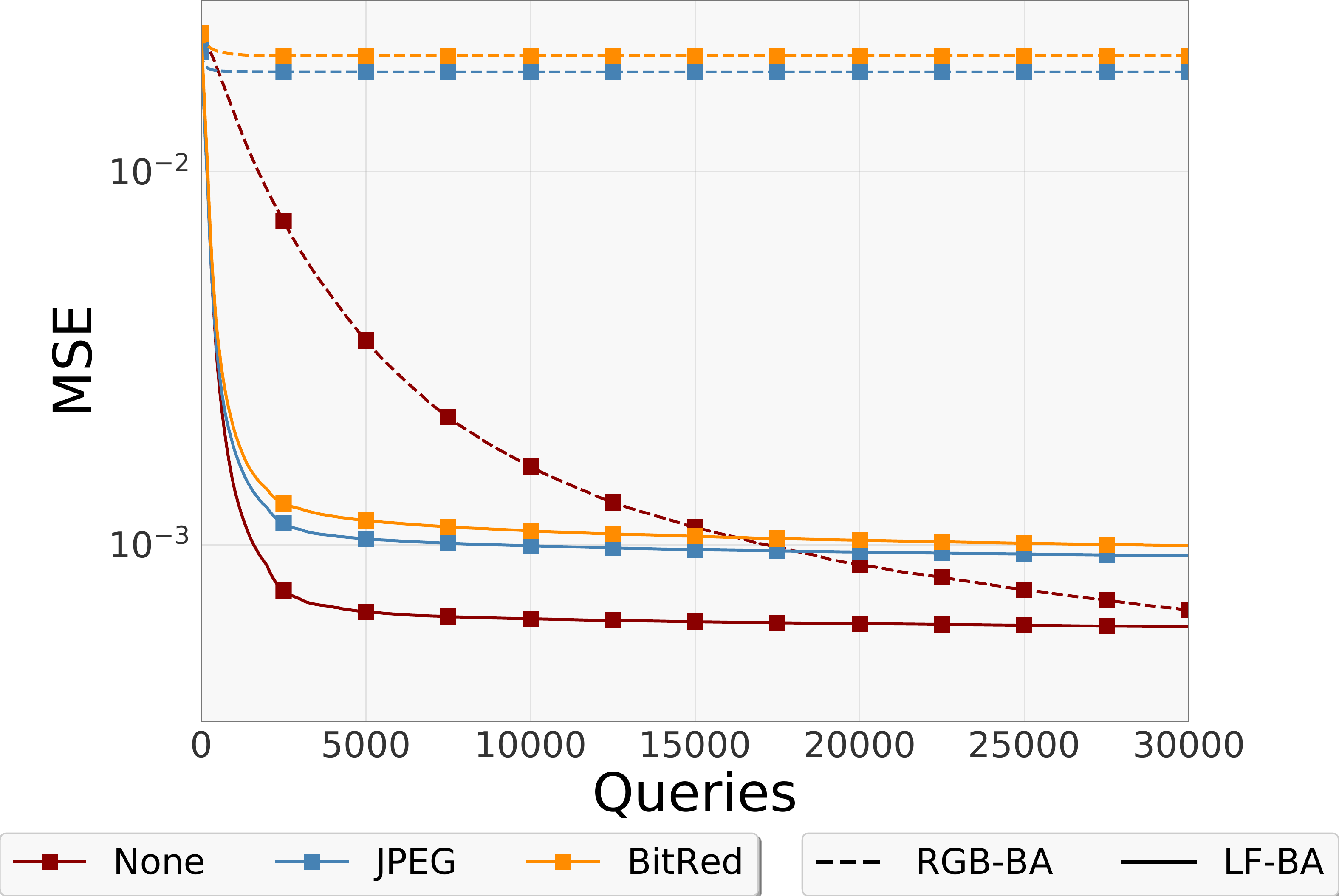}
\caption{Average (log) MSE across queries for RGB-BA and LF-BA against different image transformation defenses. Against JPEG and bit depth reduction defenses, RGB-BA fail to make progress. In contrast, LF-BA can successfully circumvent these defenses and reduce average MSE to 0.001 after $30,000$ model queries.}
\label{fig:dist_plot}
\vspace{-2ex}
\end{figure}

\paragraph{Query histogram.} \autoref{fig:queries_comparison} shows the histogram of the number of model queries required for a successful attack over 1000 sampled images. The left plot shows result for untargeted attack using RGB-BA/LF-BA. Since the boundary attack maintains an incorrectly labeled image throughout optimization while gradually reducing the perturbation norm, we define success as achieving a sufficiently low perceptibility of $<0.001$ MSE (or equivalently, an $L_2$-norm of $12.27$). The results for targeted attacks using RGB-NES/LF-NES are in the right plot. Only successful runs are included in this plot. We make several key observations:

1. The query distribution of RGB-BA (light orange) and RGB-NES (light blue) are heavy-tailed, that is, the entire range of allowed number of queries is covered, which shows that a large number of model queries is necessary for many images.

2. The histograms of LF-BA (dark red) and LF-NES (dark blue) are shifted left compared to their Gaussian-based counterparts. This demonstrates that using the low frequency noise samples consistently improves the query efficiency of the boundary attack and the NES attack. This effect is especially dramatic for LF-BA, where a large fraction of images require only roughly 1000 model queries to construct.

3. Both the median (dashed line) and mean (dotted line) query counts are significantly reduced when using LF-BA and LF-NES. In particular, LF-BA requires $1128$ median queries, an almost 4x reduction compared to the $4020$ median queries of RGB-BA. Similarly, LF-NES requires $12,444$ median queries, an approximately 2x reduction from the $22,389$ median queries of RGB-NES.

\begin{figure}[t!]
\centering
\includegraphics[width=0.85\columnwidth]{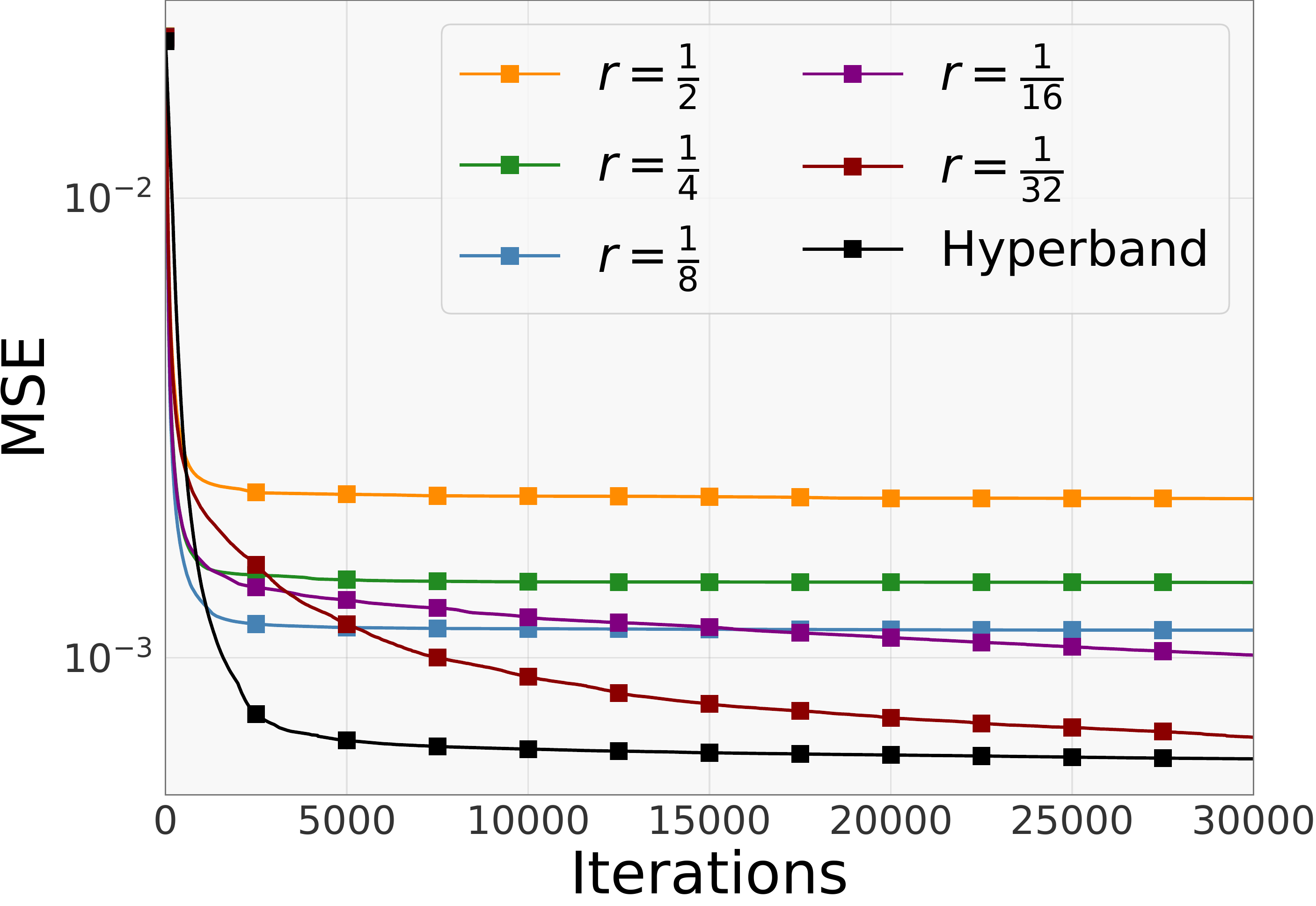}
\caption{Plot of the average perturbation MSE across iterations for LF-BA using different frequency ratios $r$. See text for details.}
\vspace{-3ex}
\label{fig:dist_plot_analysis}
\end{figure}

\begin{figure*}[t]
\centerline{\includegraphics[width=\textwidth]{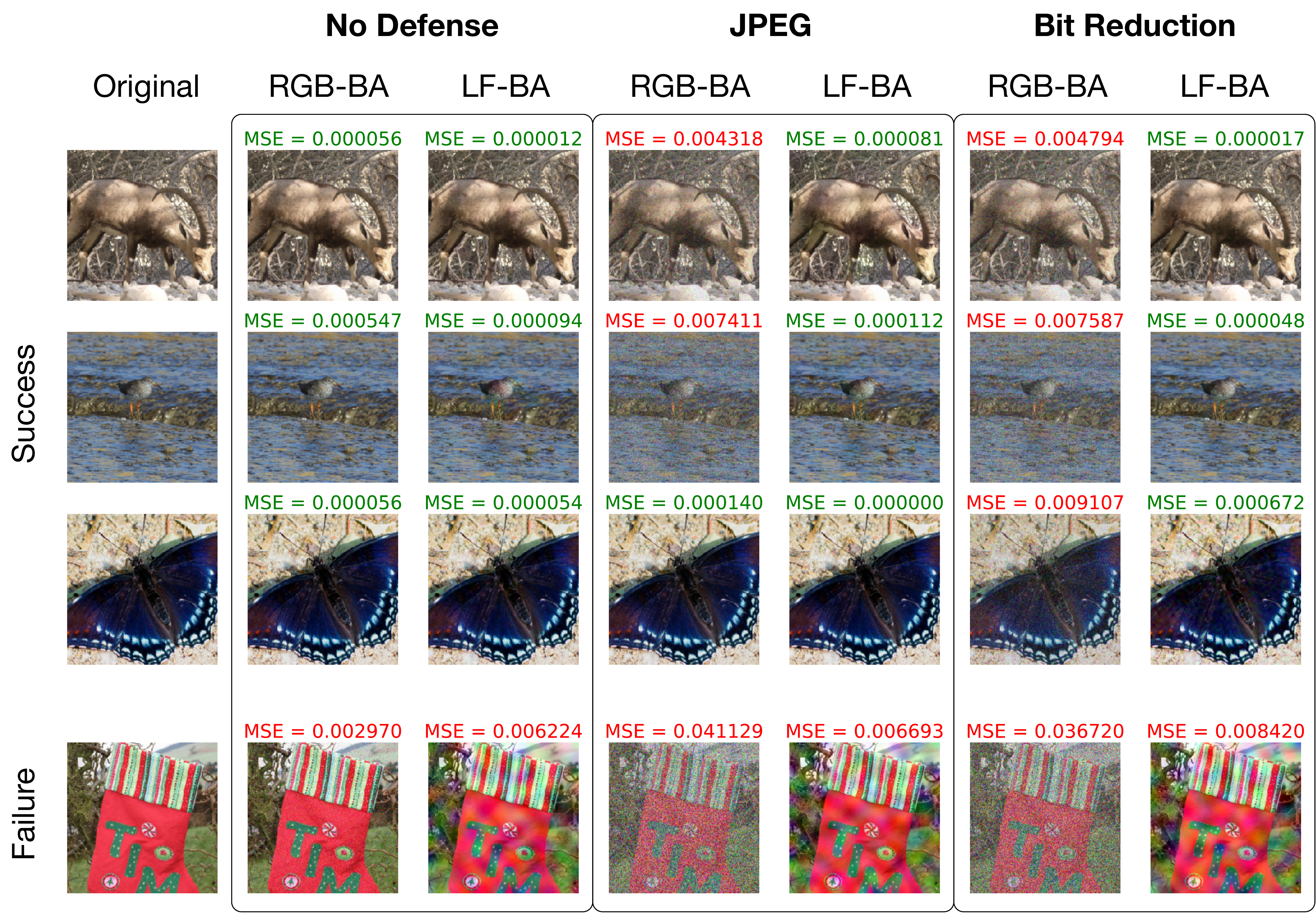}}
\caption{Image samples for attacking image transformation defenses. Perturbation MSE is truncated to 8 decimal places and images with MSE higher than 0.001 are highlighted in red. See text for details.}
\label{fig:image_samples}
\end{figure*}


\begin{table*}[t]
\centering
\resizebox{0.44\textwidth}{!}{
\begin{tabular}{cccc}
\multicolumn{4}{c}{\large \hspace{4ex} \textbf{Untargeted}} \\
\hline
\textbf{Attack} & \textbf{Average queries} & \textbf{Average $L_2$} & \textbf{Average MSE}  \\
\hline
Opt-attack & $71,100$ & 6.98 & $3.24 \times 10^{-4}$ \\
\hline
RGB-BA & $14,217$ & - & - \\
LF-BA & $2926$ & - & - \\
\hline
\end{tabular}
}
\resizebox{0.55\textwidth}{!}{
\begin{tabular}{ccccc}
\multicolumn{5}{c}{\large \hspace{3ex} \textbf{Targeted}} \\
\hline
\textbf{Attack} & \textbf{Average queries} & \textbf{Success rate} & \textbf{Average $L_2$} & \textbf{Average MSE} \\
\hline
AutoZOOM & $13,525$ & 100\% & 26.74 & $3.64 \times 10^{-3}$ \\
\hline
RGB-NES & $31,879$ & 94.7\% & 6.85 & $3.22 \times 10^{-4}$ \\
LF-NES & $17,558$ & 98.6\% & 6.92 & $3.18 \times 10^{-4}$ \\
\hline
\end{tabular}
}
\caption{Comparison of aggregate statistics on ImageNet. All statistics are averaged over 1000 random validation images. See text for details. 
\label{table:method_comparison_imagenet}}
\vspace{-1.5em}
\end{table*}

\paragraph{Selecting frequency ratio $r$.} In \autoref{fig:dist_plot_analysis} we analyze the effect of selecting the hyperparameter $r$ by either fixing it to a pre-defined value or by using Hyperband. The average MSE of adversarial perturbations constructed by LF-BA is plotted against the number of iterations. All averages are computed over the same 1000 random images from the ImageNet validation set.

At higher values of $r$, the perturbation MSE drops rapidly for the first 2500 iterations but progress stalls later on. Lower values of $r$ (e.g. purple and dark red lines) allow the attack algorithm to (relatively) slowly but eventually find an adversarial perturbation with low MSE. This plot demonstrates the need for selecting $r$ adaptively based on the image, as Hyperband (black line) selects the optimal frequency ratio to allow both rapid descent initially and continued progress later on.

\begin{figure*}[t]
\centerline{\includegraphics[width=0.8\textwidth]{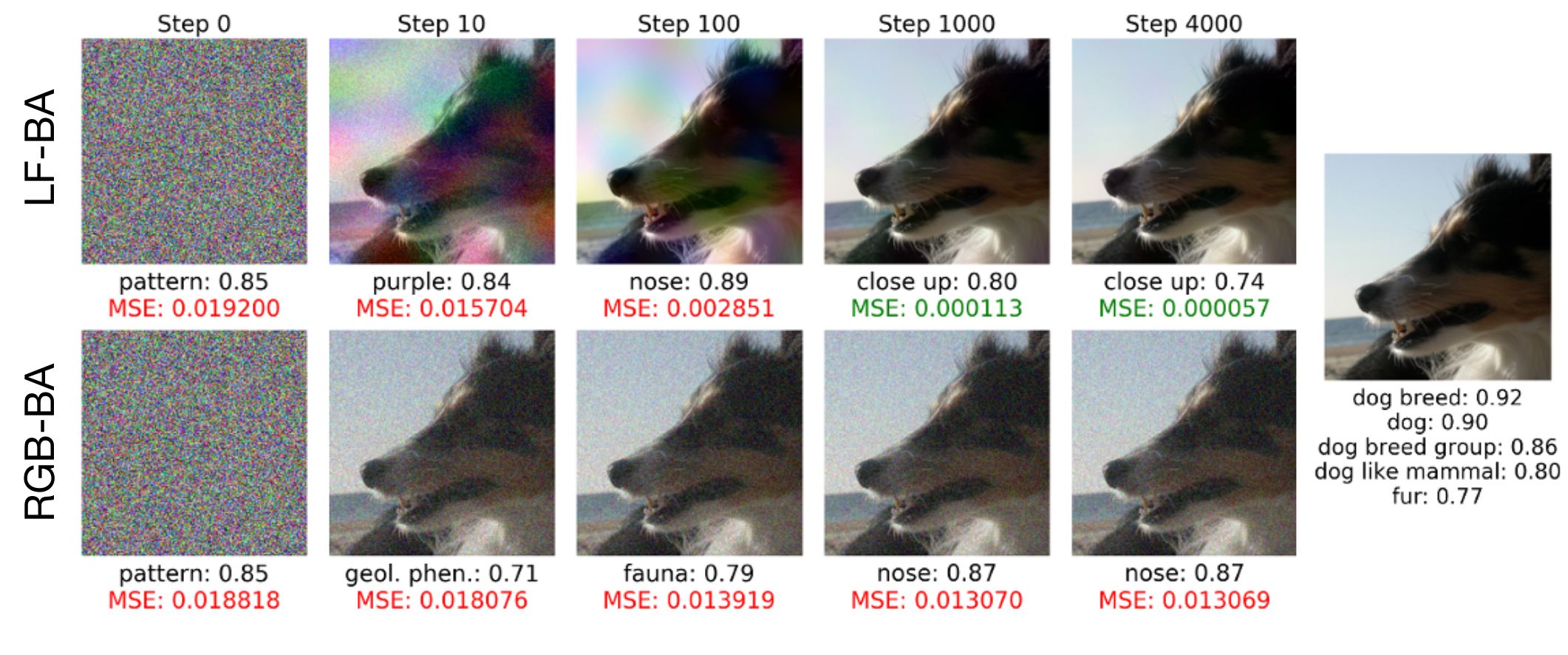}}
\vspace{-1em}
\caption{Attacking Google Cloud Vision. MSE of value higher than 0.001 is colored in red. See text for details.}
\label{fig:google_attack}
\vspace{-2ex}
\end{figure*}


\paragraph{Breaking image transformation defenses.} One common defense strategy against adversarial images is to apply a denoising transformation before feeding it into the model. This style of defense has been shown to be highly effective against transfer-based attacks \citep{guo2017countering} and have not exhibited any weakness against black-box attacks to-date. However, we suspect that low frequency perturbations can circumvent this defense since denoising transformations do not typically operate on the lower frequency spectrum.

We test our hypothesis by evaluating RGB-BA and LF-BA against the following image transformation defenses: JPEG compression \citep{dziugaite2016study} at quality level 75 and reducing bit depth \citep{xu2017feature} to 3 bits. To avoid artificially inflating success rate, we choose initial images to be correctly classified after the defensive transformation is applied.

Figure \ref{fig:dist_plot} compares the average perturbation (log) MSE across model queries for both attacks on 1000 random images across iterations. Again, we see that LF-BA (solid line) converges significantly faster than RGB-BA (dashed line) when the model is undefended (dark red lines). In fact, it reaches the same average MSE achieved by RGB-BA after $30,000$ model queries in less than $3000$ queries -- constituting an \emph{order of magnitude} reduction. When either the JPEG (blue lines) or bit depth (orange lines) reduction transformation is applied, RGB-BA fails to make any progress. This result shows that image transformation defenses are very potent against black-box attacks. On the other hand, LF-BA can circumvent these defenses consistently and reduce the average perturbation MSE to approximately $0.001$ after $30,000$ model queries. The success of LF-BA \emph{does not} rely on the knowledge of the exact transformation being applied.

Figure \ref{fig:image_samples} shows adversarially perturbed images produced when attacking different image transformation defenses. On the undefended model, there is no visible difference between the clean image and the perturbed image when attacking with either Gaussian or low frequency noise. On defended models, RGB-BA consistently fails to produce an imperceptible perturbation, while LF-BA is successful with high probability. 
Note the color patch pattern produced by LF-BA has varying frequency, which is optimally selected by Hyperband. The last image represents a failure case for both RGB-BA and LF-BA.

\paragraph{Additional baselines.} \autoref{table:method_comparison_imagenet} shows aggregate query and perturbation norm statistics for untargeted RGB-BA/LF-BA attacks and targeted RGB-NES/LF-NES attacks in comparison to two additional baselines: Opt-attack \citep{cheng2018query} and AutoZOOM \citep{tu2018autozoom}. We duplicate relevant numbers reported in the original paper for both baselines. 
Since RGB-BA/LF-BA gradually reduce perturbation magnitude at the expense of additional queries, we set a target average $L_2$-norm equal to that of Opt-attack and compare query cost. For RGB-NES/LF-NES, we fix the same maximum number of queries to $100,000$ as AutoZOOM, and compare query count and perturbation magnitude at initial success.

Note that LF-BA requires 5x fewer queries than RGB-BA and 24x fewer queries than Opt-attack to reach the same average $L_2$-norm/MSE, constituting an \emph{order of magnitude} reduction. For targeted attack, LF-NES requires only half as many queries as RGB-NES to reach the same perturbation norm while having higher success rate. Compared to AutoZOOM, LF-NES requires approximately the same number of average queries while achieving a nearly 4x reduction of perturbation $L_2$-norm.

\paragraph{Attacking Google Cloud Vision.} To demonstrate the realistic threat of low frequency perturbations, we attack Google Cloud Vision, a popular online machine learning service. The platform provides a top concept labeling functionality: when given an image, it outputs a list of top (predicted) concepts contained in the image and their associated confidence. We define a successful attack as replacing the formerly highest ranked concept with a new concept that was previously not present in the list, while obtaining an MSE $\leq 0.001$. Figure \ref{fig:google_attack} shows the progression of the boundary attack with Gaussian and low frequency noise across iterations. On the image with original top concept \emph{dog breed}, LF-BA produces an adversarial image with imperceptible difference while changing the top concept to \emph{close-up}. Even with only 1000 model queries, the adversarial perturbation is already reasonably unobtrusive. In contrast, RGB-BA could not find a sufficiently minimal perturbation within 4000 iterations (=8000 queries). Note that neither method makes use of the prediction confidence or the rank of concepts other than the top-1, contrasting with the previous known attack against this platform \citep{ilyas2018blackbox}. 

\section{DISCUSSION AND FUTURE WORK}
We have shown that adversarial attacks on images can be performed by exclusively perturbing low frequency portions of the input signal. This approach provides substantial benefits for attacks in the black-box setting and can be readily incorporated into many existing algorithms. Our follow-up work \citep{guo2019simple} that achieves state-of-the-art query efficiency using a simple coordinate descent-style attack also leverages the abundance of adversarial perturbations in the low frequency subspace.


Focusing on low frequency signal is by no means exclusively applicable to images. It is likely that similar approaches can be used to attack speech recognition systems~\citep{carlini2018audio} or time series data. Another promising future direction is to find other subspaces that may admit a higher density of adversarial perturbations. Any success in this direction can also provide us with insight into the space of adversarial examples.

\newpage
\bibliography{citations}
\bibliographystyle{aaai}

\end{document}